# On Analysis and Evaluation of Multi-Sensory Cognitive Learning of a Mathematical Topic Using Artificial Neural Networks

Fahad A. Al-Zahrani, Hassan M. Mustafa, and Ayoub Al-Hamadi

**Abstract**— This piece of research belongs to the field of educational assessment issue based upon the cognitive multimedia theory. Considering that theory; visual and auditory material should be presented simultaneously to reinforce the retention of a mathematical learned topic, a carefully computer-assisted learning (CAL) module is designed for development of a multimedia tutorial for our suggested mathematical topic. The designed CAL module is a multimedia tutorial computer package with visual and/or auditory material. So, via suggested computer package, Multi-Sensory associative memories and classical conditioning theories are practically applicable at an educational field (a children classroom). It is noticed that comparative practical results obtained are interesting for field application of CAL package with and without associated teacher's voice. Finally, the presented study highly recommends application of a novel teaching trend aiming to improve quality of children mathematical learning performance.

**Index Terms**— Associative memories, computer assisted instruction, learning, multimedia applications, neural Nets.

——————————— ◆ ———————————

## 1 INTRODUCTION

THE field of learning sciences is represented by a growing community internationally. Many educational experts now recognise that conventional ways of conceiving knowledge, educational systems and technology-mediated learning are facing increasing challenges in this time of rapid technological and social changes. Recently, that adopted trend is well supported by what had been announced (in U.S.A.) that last decade (1990-2000) called as Decade of the brain [1]. Accordingly, building up models of human brain functions considered as a recent evolutionary trend adopted by educationalists that incorporate Nero-physiology, Psychology, and Cognitive science. Consequently, Neural Networks theorists as well as neurobiologists and educationalists have focused their attention on making interdisciplinary contributions to investigate two essential brain functions (learning and memory).

More recently, Artificial Neural Networks (ANNs) combined with neuroscience considered as an interdisciplinary research direction for optimal teaching children methodology how to read, [2]. This direction motivated by a great debate given at, [3] as researches at fields of psychology and linguistic were continuously cooperating in searching for optimal methodology which supported by educational field results. Nevertheless, during last decade phonics methodology is replaced –at many schools in U.S.A.- by other guided reading methods that performed by literature based activities [4] . More recently, promising field results are obtained [4] that support optimality of phonics methodology for solving learning/teaching issue "how to read?" [2,6]. Additionally, recent mathematical modeling for phonics methodology has been presented describing memory association between two cognitive sensory auditory and visual signals at,[7]. The cognitive multimedia theory suggests that the visual and auditory material should be presented simultaneously -based upon memory association- to reinforce the retention of learned materials [8].

Herein, this cognitive theory adopted as optimal approach for improving teaching/ learning performance of a mathematical topic to children of about 11 years age. The suggested mathematical topic is teaching children algorithmic process for performing long division. Specifically for two arbitrary integer numbers chosen in a random manner (each composed of some number of digits). By detail, adopted principal algorithm for applied Computer Aided Learning (CAL) package consisted of five steps follows. Divide, Multiply, Subtract, Bring Down, and repeat (if necessary), [9]. For more details about recent view concerned with the effect of information technology computer (ITC) on mathematical education, it is advised referring to, [10]. The rest of this paper is organized as follows. At next section, a basic interactive educational model is presented with a generalized block dia-

————————————————

- *F.A. Al-Zahrani is with the Computer Engineering Department, Faculty of Computer and Information System, Umm Al-Qura University, Makkah Saudi Arabia.*
- *H.M. Mustafa is with Computer Engineering Department, Faculty of Engineering, Albaha University. On leave from the Educational Technology Department-Faculty of Specified Education-Banha University, Egypt .*
- *A. Al-Hamadi is with the Institute for Electronics, Signal Processing and Communications (IESK) Otto-von-Guericke-University, Magdeburg.*

gram. A generalized ANN learning / teaching mode is given at third section. At the fourth section, obtained results are given after application of suggested CAL package at the case study. Additionally, obtained simulation results are presented at fifth section. Finally, at the sixth section, some interesting conclusions, suggestions for future work are presented .The algorithm, for simplified flow chart of adopted CAL package is given at an Appendix.

## 2. BASIC LEARNING/TEACHING MODEL

Generally, practical performing of learning process - from neurophysiologic P.O.V. - utilises two basic and essential cognitive functions. Both functions are required to perform efficiently learning / teaching interactive process in accordance with behaviourism, [11-13], as follows. Firstly, pattern classification/recognition function based on visual/audible interactive signals stimulated by CAL packages. Secondly, associative memory function is used which is originally based on classical conditioning motivated by Hebbian learning rule. Referring to Fig.1 shown in below, the illustrated teaching model is well qualified to perform simulation of above mentioned brain functions. Inputs to the neural network learning model at that Figure, are provided by environmental stimuli (unsupervised learning).The correction signal for the case of learning with a teacher is given by responses outputs of the model will be evaluated by either the environmental conditions (unsupervised learning) or by the teacher. Finally, the tutor plays a role in improving the input data (stimulating learning pattern), by reducing noise and redundancy of model pattern input. That is according to tutor's experience, he provides the model with clear data by maximizing its signal to noise ratio. However, that is not our case which is based upon unsupervised Hebbian self-organized (autonomous) learning,[14] .Details of mathematical formulation describing memory association between auditory and visual signals are given at,[7].

## 3. A GENERALIZED ANN LEARNING/TEACHING MODEL

The presented model given at Fig.2 in below generally simulates two diverse learning paradigms. It presents realistically both paradigms : by interactive learning / teaching process, as well as other self organized (autonomous) learning  By some details , firstly is  concerned with classical (supervised by tutor) learning observed at our classrooms (face to face tutoring) . Accordingly, this paradigm proceeds interactively via bidirectional communication process between teacher and his learner(s) [14]. However, secondly other learning paradigm performs self-organized (autonomously unsupervised) tutoring process [15].

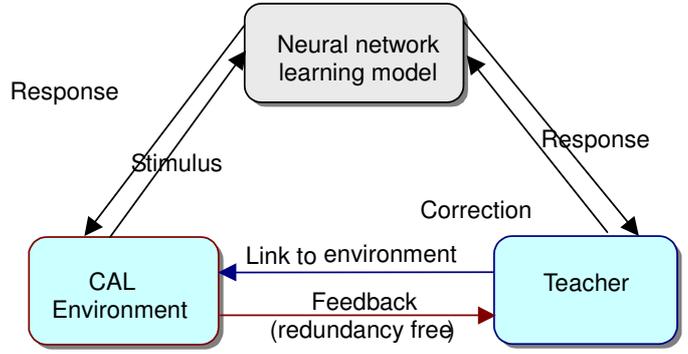

Fig.1: Illustrates a general view for interactive educational process, adapted from [6].

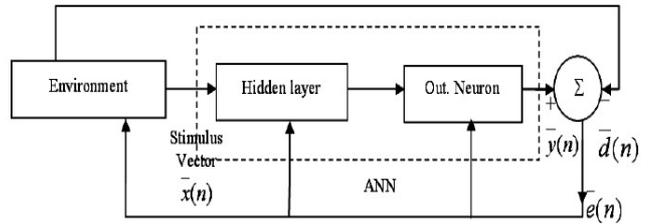

Fig.2: Generalized ANN block diagram simulating two diverse learning paradigms, adapted from [7].

The error vector at any time instant (n) observed during learning processes is given by:

$$\overline{e}(n) = \overline{y}(n) - \overline{d}(n) \qquad (1)$$

Whereby $\overline{e}(n)$ is the error correcting signal controlling adaptively the learning process and $\overline{y}(n)$ is the output signal of the model. $\overline{d}(n)$ is the desired numeric value(s). Referring to above Fig.1; following four equations are deduced:

$$V_k(n) = X_j(n) W_{kj}^T(n) \qquad (2)$$

$$Y_k(n) = \varphi(V_k(n)) = (1 - e^{-\lambda V_k(n)}) / (1 + e^{-\lambda V_k(n)}) \qquad (3)$$

$$e_k(n) = |d_k(n) - y_k(n)| \qquad (4)$$

$$W_{kj}(n+1) = W_{kj}(n) + \Delta W_{kj}(n) \qquad (5)$$

Whereby $X$ is input vector and $W$ is the weight vector. $\varphi$ is the activation function. $Y$ is the output. $e_k$ is the error value and $d_k$ is the desired output. Noting that $\Delta W_{kj}(n)$ the dynamical change of weight vector value. Above four equations are commonly applied for both learning paradigms: supervised (interactive learning with a tutor), and unsupervised (learning though students' self-study). The dynamical changes of weight vector value specifically for supervised phase is given by equation:

$$\Delta W_{ki}(n) = \eta e_k(n) X_i(n) \qquad (6)$$

Where η is the learning rate value during learning process for both learning paradigms. However, for unsu-pervised paradigm, dynamical change of weight vector value is given by equation:

$$\Delta W_{ki}(n) = \eta Y_k(n) X_i(n) \qquad (7)$$

Noting that $e_k(n)$ in (6) is substituted by $y_k(n)$ at any arbitrary time instant (n) during learning process.

## 4. CASE STUDY RESULTS

The results obtained after performing practical experimental work in classroom (case study) are shown in graphical and tabulated forms at the two subsequent subsections 4.1., and 4.2. respectively.

### 4.1. Graphical Obtained Results

The two figures Fig.3 & Fig.4 shown in below illustrate two bi-comparative tutoring performances. Noting that both tutoring computer programs (without audio, and with audio tools), are having superiority with respect to the classical teaching approach.

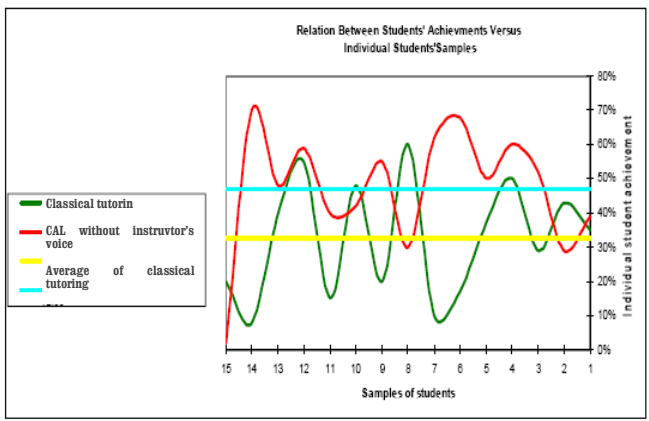

Fig.3: An illustrative bi-comparative graphical figure for classical teaching versus tutoring using computer program without audio (instructor's voice)

### 4.2. Tabulated Obtained Results:

A learning style is a relatively stable and consistent set of strategies that an individual prefers to use when engaged in learning [16-17]. Herein, our practical application (case study) adopts one of these strategies namely acquiring learning information through two sensory organs (student eyes and ears). In other words, seen and heard (visual and audible) interactive signals are acquired by student's sensory organs either through his teacher or considering CAL packages (with or without teacher's voice). Practically, children are classified in three groups in according to their diverse learning styles (preferences).

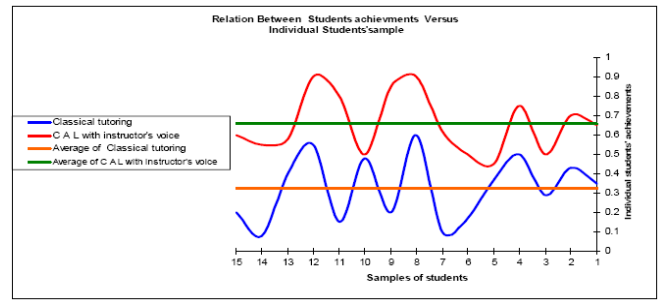

Fig.4: An illustrative bi-comparative graphical figure for classical teaching versus tutoring using computer program with audio (instructor's voice)

The two tables (Table.1 & Table.2) given in below illustrate obtained practical results after performing three different learning experiments. At table.1, illustrated results are classified in accordance with different students' learning styles following three teaching methodologies. Firstly, the classical learning style is carried out by students-teacher interactive in the classroom. Secondly, learning is taken place using a suggested software learning package without teacher's voice association. The last experiment is carried out using CAL package that is associated with teacher's voice. This table gives children's achievements (obtained marks) considering that maximum mark is 100. The statistical analysis of all three experimental marking results is given in details at Table.2 shown in below.

Table.1
Illustrates children's marks after performing three educational experiments

| | | | | | | | | | | | | | | | |
|---|---|---|---|---|---|---|---|---|---|---|---|---|---|---|---|
| Classical Learning | 35 | 43 | 29 | 50 | 37 | 17 | 10 | 60 | 20 | 48 | 15 | 55 | 40 | 8 | 20 |
| CAL without Voice | 39 | 29 | 52 | 60 | 50 | 68 | 62 | 30 | 55 | 42 | 40 | 59 | 48 | 70 | 2 |
| CAL with Voice | 65 | 70 | 50 | 75 | 45 | 50 | 62 | 90 | 85 | 50 | 80 | 90 | 58 | 55 | 60 |

Table.2: Illustrates statistical analysis of above obtained children's marks

| Teaching Methodology | Children's average Achievement score (M) | Variance $\sigma$ | Standard deviation $\sqrt{\sigma}$ | Coefficient of variation $\rho = \sqrt{\sigma}/M$ | Improvement of teaching Quality |
|---|---|---|---|---|---|
| Classical | 32.46 | 265.32 | 16.28 | 0.50 | - |
| CAL (without tutor's voice) | 46.80 | 297.49 | 17.24 | 0.36 | 44.1% |
| CAL (with tutor's voice) | 64.33 | 283.42 | 16.83 | 0.26 | 98.2% |

## 5. SIMULATION RESULTS

At this section realistic simulation results are introduced in two tabulated as well as graphical forms. Interestingly, those presented simulation results are in well agreement with practically obtained results shown at subsections 4.1 and 4.2 given in the above.

The suggested ANN model adapted from realistic learning simulation model given at [6] with considering various learning rate values. It is worthy to note that learning rate value associated to CAL with teacher's voice proved to be higher than CAL without voice. Simulation curves at Fig.5 illustrate statistical comparison for two learning processes with two different learning rates. The lower learning rate ($\eta = 0.1$) may be relevant for simulating classical learning process. However, higher learning rate ($\eta = 0.5$) could be analogously considered to indicate (approximately) the case of CAL process applied without teacher's voice.

Table 3: Illustration of simulation results for different learning rate values.

| Learning Rate value | Children's average Achievement score (M) | Variance $\sigma$ | Standard deviation $\sqrt{\sigma}$ | Coefficient of variation $\rho = \sqrt{\sigma}/M$ | Improvement of teaching Quality |
|---|---|---|---|---|---|
| $\eta=0.1$ | 42 | 428.5 | 20.7 | 0.61 | - |
| $\eta=0.5$ | 64 | 918.1 | 30.3 | 0.47 | 66% |

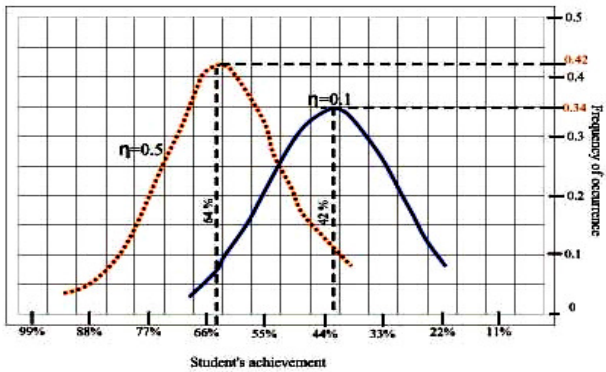

Fig.5: Illustrates Simulation results presented by statistical distribution for children's achievements versus the frequency of occurrence for various achievements values, at different learning rate values ( = 0.1 & = 0.5).

## 6. CONCLUSION

Interestingly, it is clear that both practical obtained as well as realistic simulation results are in agreement with each other as illustrated in the above fourth and fifth sections respectively. Additionally, some other results support our presented work issue as at [18-20]. That likewise, the cognitive multimedia theory suggests that the visual and auditory material should be presented simultaneously to allow the individual to make connections between the types of material rather than presenting the material successively.

Referring to [18], Lindstrom found that participants could only remember 20% of the total materials when they were presented with visual material only, 40% when they were presented with both visual and auditory material, and about 75% when the visual and auditory material were presented simultaneously. [8] . Moreover, detail evaluation and analysis for presented work resulted in the following five interesting conclusive remarks:

1- Evaluation of any CAL package quality is measured after analysis of educational field results. So, above suggested strategy provides specialists in educational field with fair unbiased judgment for any CAL package. That is by fair comparison of statistical analysis parameters after obtained simulation results of children's achievements with practical natural analysis of obtained individual differences results. Consequently, more effective modification for any CAL package could be fulfilled if and only if analysis of statistical measured practical results in advised an additive value for learning effectiveness parameters (better quality).

2- The above obtained results after practical application of our multimedia CAL (case study) are depending only upon two Multi-Sensory sensory cognitive systems (visual and/or audible) while performing learning process. By future application of three Multi-Sensory Cognitive Learning, CAL systems based on virtual reality technique, will add one more educational sensory system (tactile system) contributed in learning process. Accordingly, application of such three sensory virtual reality learning ( including tactile, visual, and audible) system more promising in development of an additive value for learning / teaching effectiveness .

3-Recently, timely considered learning phenomenon, and based on cognitive psychology. That is considerate computing applied to modify learning systems. These systems have to monitor interruption phenomena carried out by students following computer screen (VDU) activities [21]. Consequently, sensing attention mainly exculpated to perform the function of future considerate learning computer systems [22]. That equipped by attentive appliances which are responsible for gaze detection function during learning by video conference systems [23].

4-For future modification of suggested CAL package is measurement of time learning parameter will be promising for more elaborate measurement of learning performance in practical educational field (classroom) application. This parameter is recommended for educational field practice, [24] as well as for very recently suggested measuring of e-learning systems performance [25].

5-Finally, for future extension of presented research work, it is highly recommended to consider more elaborate investigational analysis and evaluations for other behavioral learning phenomena observed at educational field (such as learning creativity, improvement of learning performance, learning styles,……etc.) using ANNs modeling. As consequence of all given in above, it is worthy to recommend realistic implementation of ANNs models , to be applicable for solving educational phenomena issues related to cognitive styles observed at educational phenomena and/or activities as that introduced at [26].

**Fahad Ahmed Al-Zahrani** received the B.Sc. degree in computer engineering from Umm Al-Qura University, Makkah, Saudi Arabia, in 1996 and the M.S. in computer engineering from Florida Institute of Technology, Melbourne, Florida, in 2000. He received the Ph.D. degree in computer engineering from Colorado state university, Fort Collins, Colorado, in 2005.

Since November 2005, he has been an Assistance Professor in the Department of Computer Engineering, Umm Al-Qura University, Makkah, Saudi Arabia. His research interests include communication networks, all-optical networks design and performance evaluation, fiber optic communication, and network security. He is a member of both IEEE Computer Society and IEEE Communication Society. He is a member of the international Society for Optical Engineering. In addition, he is a member of Optical Society of America.

**Hassan Mohammed Hassan**: was born in Cairo, on October 1947. He received his B.Sc. degree in Electrical Engineering and his M.Sc. degree from Military Technical College (M.T.C.) Cairo-Egypt in 1970, and 1983 respectively.  He received his Ph. D. degree in Computer Engineering and Systems in 1996 from Ain Shams University - Faculty of Engineering, Cairo –Egypt.   Currently  he is Associate Professor of Computer Science& Information with IT& Computer Dept. Arab Open University (Kingdom of Saudi Arabia Branch,) He is a member of some Scientific Societies associated with Education & computer ,and Communication technologies such as IIIS (International Institute of Informatics and Systemics) . Moreover, he is a member at IAOE International Association of Online Engineering. Recently, he has been appointed as a member of technical comity at IASTED organization during (2009-2011). Additionally, he  has been invited for attending  as German DAAD fellowship  for two research programs during academic year 2008/2009  at Institute for Electronics, Signal Processing and Communications (IESK) Otto-von- Guericke University Magdeburg. His interest fields of research are Artificial Neural Networks, Natural Inspired Computations, and their applications for modeling of Communication systems and evaluation of learning processes /phenomena. He is author of more than 50 published/accepted papers technical articles at international specialized conferences and journals. That is during time period from 1982 till 2009.

**Ayoub Al-Hamadi** was born in Taiz, Yemen in 1970 .He received his Masters Degree in Electrical Engineering & Information Technology in 1997 and his PhD. in Technical Computer Science at the Otto-von-Guericke-University of Magdeburg, Germany in 2001. Since 2003 he has been Junior-Research-Group-Leader at the Institute for Electronics, Signal Processing and Communications at the Otto-von-Guericke-University Magdeburg. In 2008 he became Professor of Neuro-Information Technology at the Otto-von-Guericke University Magdeburg. Prof. Dr.-Ing. Al-Hamadi is the author of more than 90 articles in peer-reviewed international journals and conferences. His Research Fields: Image Processing and Pattern recognition, Human Computer Interaction, Information Technology, and Artificial Intelligence.

# APPENDIX

A simplified macro-flowchart of adopted CAL package describing briefly algorithmic steps for performing long division process by referring to [9].

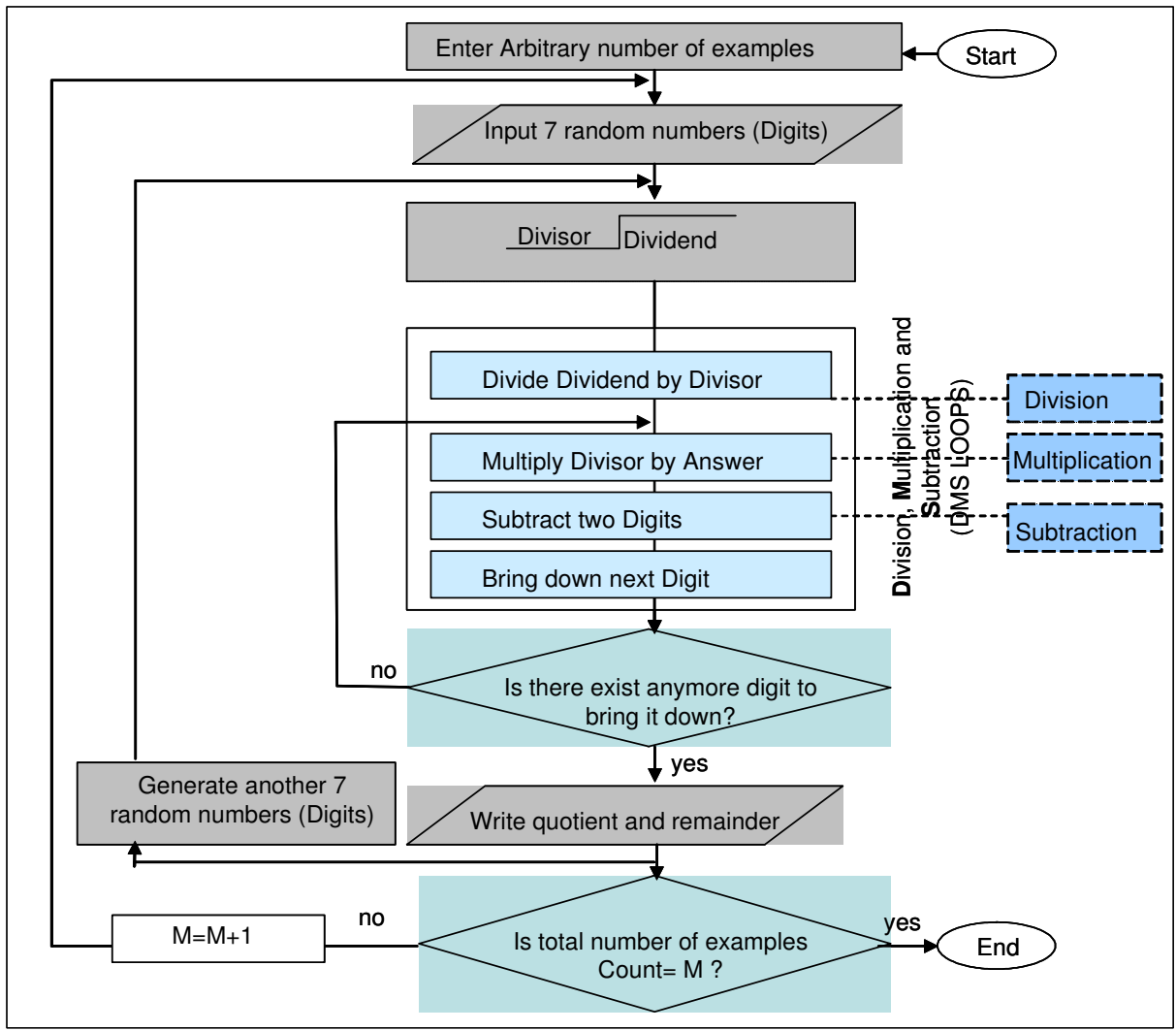